# Gait Pattern Recognition Using Accelerometers

Vahid Alizadeh
University of Michigan
Alizadeh@umich.edu

## Abstract
**Motion ability is one of the most important human properties, including gait as a basis of human transitional movement. Gait, as a biometric for recognizing human identities, can be non-intrusively captured signals using wearable or portable smart devices. In this study gait patterns is collected using a wireless platform of two sensors located at chest and right ankle of the subjects. Then the raw data has undergone some preprocessing methods and segmented into 5 seconds windows. Some time and frequency domain features is extracted and the performance evaluated by 5 different classifiers. Decision Tree (with all features) and K-Nearest Neighbors (with 10 selected features) classifiers reached 99.4% and 100% respectively.**

## 1. INTRODUCTION

There are different biometric approaches for identity recognition and one of the most prominent ones are through image processing of human face [1]. Development of microelectromechanical systems (MEMS) in recent years has influenced in many research areas. One of the most important group of these systems are inertial sensors, consists of accelerometer and gyroscope. One area that absorbed many attentions recently is gait analysis. Motion ability including gait is one of the most important traits of human since it has considerable influence on quality of life. Gait analysis approaches are applicable in security or healthcare areas. On the other hand, with the fast development of ubiquitous devices in which inertial sensors has been integrated, the applications of motion analysis methods has been extended to activity recognition and sports [2].

## 2. WEARABLE SENSOR PLATFORM

### 1.1 Participants and Platform

The data collection phase of this study was conducted at the University of Michigan, WSSP lab, to collect the data for gait pattern recognition. A total of 4 subjects participated in data collection procedure. A set of inertial measurement units was attached to each subject. Since two different research combined together, other sensors such as EEG, ECG, and GSR was used too. Shimmer wireless sensor platform [3] is used for gait recognition purpose. Shimmer is small and robust wearable wireless sensor with a 24MHz CPU. It contains inertial sensing via accelerometer and gyroscope with selectable range. The data was recorded synchronously from two sensors located on right side of right ankle and center of chest of each subject. The sampling rate is approximately 50Hz. An exclusive Java application developed by WSSP lab is used to collect synchronized data of all sensors and annotate them with desired label.

### 2.1 Procedure

Each subject was asked to walk naturally from the lab to a specific point outside of the building. Therefor the path was approximately equal for all the subject. The subjects should pass from two doors during their walking and take one left turning. One important consideration in this experiment was type of the floor since it plays a significant role on human gait patterns [4]. The floor during the selected path was different between inside and outside of the building. Another consideration was the subject's outfit. Two different session of data collection were conducted with at least one week interval between them and the subjects were asked to wear different outfits especially shoes. The reason was to consider the effect of the outfit and subject's mood on their gait patterns.

## 3. DATA ANALYSIS

Due to the nature of raw data output of two sensor nodes, it is impossible to use them to classify subjects. Fig shows 3-dimensional raw accelerometer data for the ankle node. Therefore several features were extracted from these raw signals and used as input to the classifiers. The features were chosen based on the application and the experiment platform (Sensor network locations on the subject's body). In this way, the complexity of the raw data is reduced and some meaningful information is extracted from the raw data. My data analysis recipe includes pre-processing, synchronization, calibration, segmentation, feature extraction, and classification. The overview of data analysis procedure has been shown in Figure 1. Each steps is described in this section and the results will be provided in the next section. MATLAB software is used for data analysis purposes.

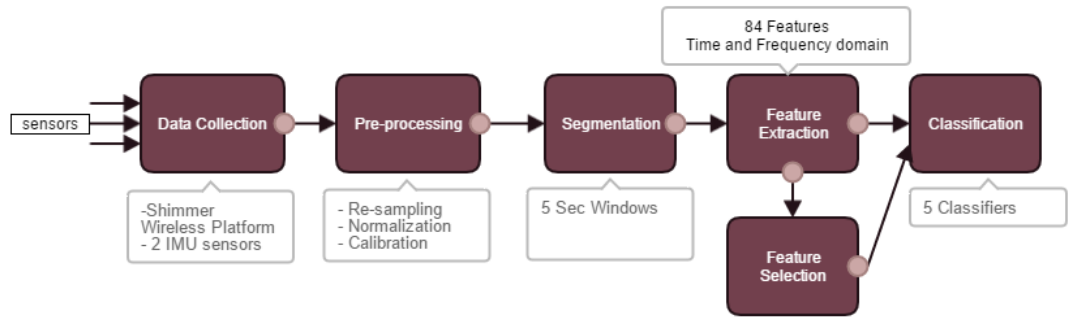

*Figure 1 Overview of data analysis procedure*

### 3.1 Pre-Processing

First the raw output of signal related to the experiment (activity) is segmented. Then null values in data matrix caused by different sampling rate of sensors are find and replaced using linear interpolation. After that the raw signal get calibrated. The purpose of the calibration is to convert the instrument readings to the units of interest and to eliminate or reduce bias in an instrument's readings.

The sampling rate of the raw accelerometer data is not exactly at 50 Hz and can be slightly change during data collection due to the hardware and wireless limitation. This cause a non-uniformly sampled data which can impact the results. Therefore, a resampling algorithm is implemented to have a uniformly sampled data at 50 Hz. This results to better feature identification and improves the classification performance [5].

The final raw signal is not zero-mean. Subtracting the mean from the data for each axis signal will remove any constant effects, such as gravity.

### 4.1 Segmentation

Before extracting features, the raw signals chunked using a sliding window. The window size of 5 seconds is chosen without any overlap. This length is enough to extract stable and consistent features to feed to the classifiers. Then feature extraction algorithm applied to each window to generate a feature vector (sample). These samples are labeled with the specific subjects.

### 5.1 Feature Extraction

A total of 84 features were extracted from each window of signal. These features listed as follows: Statistical features such as Min, Max, standard deviation; the mean of magnitude. Magnitude of the signal can be calculated by equation (1); the squared sum of magnitude below 25 and 75 percentile; the median frequency such that equation (2) is satisfied.

$$Mag(t) = \sqrt{x^2(t) + y^2(t) + z^2(t)} \qquad (1)$$

$$\sum_0^{MF} PSD(f) = \frac{1}{2}\sum PSD(f) \qquad (2)$$

Peak frequency in spectrum below $f_s$ Hz; integral of spectrum from 0 to $f_s$ Hz; number of peaks in spectrum below $f_s$ Hz; the coordinates of the two first peaks. $f_s$ is considered equal to 5Hz. List of all features and their descriptions is referenced in Table 1 [6][7][8].

*Table 1 List of all features extracted from each window of the signal*

| NO. | NAME | DESCRIPTION |
| --- | --- | --- |
| 1 | Max | *The maximum value of the signal* |
| 2 | Min | *The minimum value of the signal* |
| 3 | Mean | *The average value of the signal* |
| 4 | Median | *The median value of the signal* |
| 5 | StD | *Measures the amount of variation from the average value* |
| 6 | MF | *Median Frequency* |
| 7 | PeakX1 | *Abscise of the first peak* |
| 8 | PeakX2 | *Abscise of the second peak* |
| 9 | PeakY1 | *Ordinate of the first peak* |
| 10 | PeakY2 | *Ordinate of the second peak* |
| 11 | PeakFreq | *Peak frequency in spectrum below 5 Hz* |
| 12 | NumPeak | *Number of peaks in spectrum below 5 Hz* |
| 13 | IntegSpec | *Integral of spectrum from 0 to 5 Hz* |
| 14 | AM | *The average magnitude of all the sample points* |
| 15 | SqSum25 | *The squared sum of magnitude below 25 percentile* |
| 16 | SqSum75 | *The squared sum of magnitude below 75 percentile* |

Features 1 to 13 were computed for each three axes while features 14 to 16 obtained from all three axes of both sensor nodes. Therefore, a total of 84 features were extracted from each window of the signal.

In order to eliminate the bias in each feature across all the subjects, the values of all features is normalized to confine the range of values to be between [0,1]. Before normalization, outliers should be removed since they scale the data to a small range.

### 6.1 Classification

Five Classification methods were used to classify gait patterns: Decision tree, Linear Discriminant, K-Nearest Neighbor, Support Vector Machine, and Naïve Bayes.

Decision tree classifier [9] is a method commonly used in machine learning and data mining and uses a decision tree as a predictive model which maps observations about an item to conclusions about the item's target value. A decision tree is a flow-chart-like structure, where each internal (non-leaf) node denotes a test on an attribute, each branch represents the outcome of a test, and each leaf (or terminal) node holds a class label. The topmost node in a tree is the root node. There are different algorithms for constructing decision trees in which different metrics apply to provide a measure of the quality of the split.

Linear discriminant analysis (LDA) [10] is a method used in statistics, pattern recognition and machine learning to find a linear combination of features that characterizes or separates two or more classes of objects or events. The resulting combination may be used as a linear classifier, or, more commonly, for dimensionality reduction before later classification.

Nearest neighbor classifiers [11] are a class of non-parametric methods used in pattern recognition. The method classifies objects based on closest training sample point in the feature space. The k-nearest neighbor classifier assigns a point to a particular class based on a majority vote among the classes of the k nearest training points.

A Support Vector Machine (SVM) [12] is a supervised learning model which classifies data by finding the best hyperplane that separates all data points of one class from those of the other class. The best hyperplane for an SVM means the one with the largest margin between the two classes. Margin means the maximal width of the slab parallel to the hyperplane that has no interior data points. The support vectors are the data points that are closest to the separating hyperplane; these points are on the boundary of the slab.

The accuracy of classifiers were measured using K-fold cross validation. In this method, the data set is divided into k subsets, and the holdout method is repeated k times. Each time, one of the k subsets is used as the test set and the other k-1 subsets are put together to form a training set. Then the average error across all k trials is computed. The advantage of this method is that it matters less how the data gets divided. Every data point gets to be in a test set exactly once, and gets to be in a training set k-1 times. The variance of the resulting estimate is reduced as k is increased. The disadvantage of this method is that the training algorithm has to be rerun from scratch k times, which means it takes k times as much computation to make an evaluation.

## 4. RESULTS AND DISCUSSION

First all features were used to train classifiers. The classification accuracy were assessed using 5-fold cross validation method. Table 2 represents the detailed results of classification.

*Table 2 Classification results using all features assessed by 5-fold cross validation*

|  | **Decision Tree** | **Linear Discriminant** | **Nearest Neighbor** | **Support Vector Machine (OvO)** | **Naïve Bayse** |
|---|---|---|---|---|---|
| **Accuracy** | 99.4% | 71.4% | 73.3% | 71.9% | 86.24% |
| **Speed (s)** | 21 | 8 | 28 | 41 | 14 |

It can be observed from the above table that Decision Tree classifier has the best correct classification rate. Confusion Matrix, true positive rate, and false negative rate is shown in Figure 2.

Since there are too many features, in the next step feature selection has been tried. Feature selection is different from dimensionality reduction. Both methods seek to reduce the number of attributes in the dataset, but a dimensionality reduction method do so by creating new combinations of attributes, whereas feature selection methods include and exclude attributes present in the data without changing them. Keeping irrelevant features can result in overfitting. Decision tree algorithms try to make optimal spits in feature values. Those features that are more correlated with the prediction are split on first. Deeper in the tree less relevant and irrelevant features are used to make prediction decisions that may only be beneficial by chance in the training dataset. This overfitting of the training data can negatively affect the modeling power of the method and ruin the predictive accuracy. Therefore, it is important to remove redundant and irrelevant features. For this purpose, WEKA tool is used with these options: the attribute evaluator as *CfsSubsetEval* (Values subsets that correlate highly with the class value and low correlation with each other) and search method as *BestFirst* (Uses a best-first search strategy to navigate attribute subsets). Finally, 10 features has been selected: Ankle Node (PeakFreq_X, IntegSpec_X, PeakFreq_Z, IntegSpex_Z, NumPeak_Z, PeakX1_X, PeakX2_X, PeakX2_Y, PeakX1_Z) Chest Node (PeakX1_X). Then, the same classifiers is applied using just the selected features and the result is shown in Table 3.

*Table 3 Classification results using 10 selected features by Best First algorithm assessed by 5-fold cross validation*

|  | **Decision Tree** | **Linear Discriminant** | **Nearest Neighbor** | **Support Vector Machine (OvO)** | **Naïve Bayse** |
|---|---|---|---|---|---|
| **Accuracy** | 98.5%| | 83.9% | 100% | 86.90% | 85.90% |
| **Speed (s)** | 7 | 5 | 9 | 86 | 13 |

As it can be seen, utilizing the optimal feature set will lead to improving the performance of linear Discriminant, Nearest Neighbor and SVM in comparison to using all features and especially 100 percent correct classification rate for 1-Nearest Neighbor classifier. Confusion Matrix, true positive rate, and false negative rate for this classifier is shown in Figure 2. Note that as it was expected, the speed of building classifiers models and evaluation phase decreased considerably after dimensionality reduction of feature space.

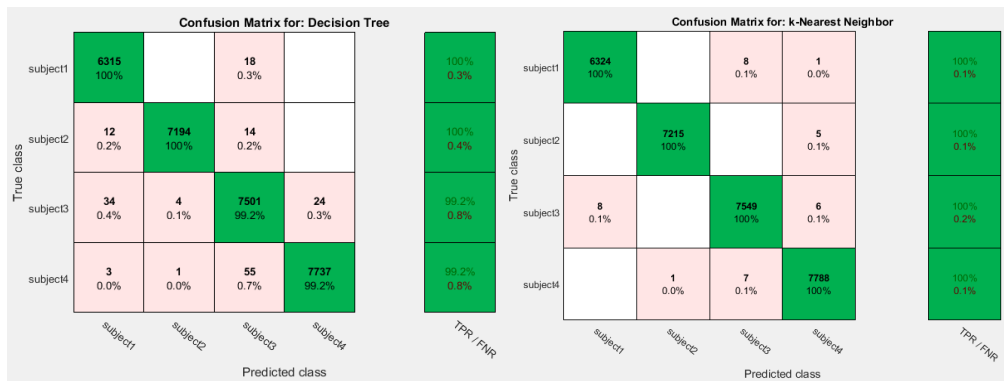

*Figure 2 Confusion Matrix for Decision Tree classifier with all features (left) and 1-Nearest Neighbor with selected features (right)*

Although a few subjects is studied in this study, the results are still promising showing that the extracted features have enough discrimination capability to use in gait pattern recognition applications.

## 5. CONCLUSIONS

Considering the fact that each individual has a unique way of walking, inertial sensors such as accelerometers can be used for gait recognition where assessed gait can be interpreted as a biometric feature. In this study, gait data is collected using Shimmer wireless platform. The experiment procedure designed in a generalized manner to consider non-straight walking path, different floor type, different outfit and mood. All these factors affects the subject's gait pattern based on previous researchers' works in this area. Then, the raw data is processed and some feature is extracted. Finally, the performance of different classifiers were assessed.

Some methods and concepts from Data Mining course were employed such as data pre-processing, data interpolation, resampling, normalization, decision tree and naïve Bayes classifiers, and feature selection. The shortcoming of this study was the limited number of subjects, but the fact that dealing with data collection phase of this study instead of using other's data sets, which consists a large portion of this work, and applying the methods were taught in Data Mining course, definitely provides an invaluable experience for the researcher. The larger data set would provide a better groundwork to assess the feature's effectiveness.

The future works is to consider the impact of segmentation window size and studying the effect of an object carried by the subject on his/her gait patterns.

Furthermore, designing a more generalized model which addresses the sensors' dislocation and disorientation would be a promising path to continue this work on.

## REFERENCES

[1] V. Alizadeh, S. RayatDoost, and E. Arbabi, "Effect of different partitioning strategies of face imprint on thermal face recognition," in *2014 22nd Iranian Conference on Electrical Engineering (ICEE)*, 2014, pp. 1108–1112.

[2] S. Sprager and M. B. Juric, "Inertial Sensor-Based Gait Recognition: A Review," *Sensors*, vol. 15, no. 9, pp. 22089–22127, Sep. 2015.

[3] "Shimmer, Wearable Sensor Technology," 2015. [Online]. Available: http://www.shimmersensing.com/. [Accessed: 01-Jan-2015].

[4] A. Muro-de-la-Herran, B. García-Zapirain, and A. Méndez-Zorrilla, "Gait Analysis Methods: An Overview of Wearable and Non-Wearable Systems, Highlighting Clinical Applications," *Sensors*, vol. 14, no. 2, pp. 3362–3394, 2014.

[5] C. E. Lunneborg, *Data analysis by resampling: Concepts and applications*. Brooks/Cole, 2000.

[6] B. Caby, S. Kieffer, M. de Saint Hubert, G. Cremer, and B. Macq, "Feature extraction and selection for objective gait analysis and fall risk assessment by accelerometry," *Biomed. Eng. Online*, vol. 10, no. 1, p. 1, 2011.

[7] J. Frank, S. Mannor, J. Pineau, and D. Precup, "Time Series Analysis Using Geometric Template Matching," *IEEE Trans. Pattern Anal. Mach. Intell.*, vol. 35, no. 3, pp. 1–1, 2012.

[8] N. A. Y. Ma, S. Henry, A. Kierlanczyk, M. Sarrafzadeh, J. Caprioli, K. Nouri-Mahdavi, H. Ghasemzadeh, "Investigation of Gait Characteristics in Glaucoma Patients with a Shoe-Integrated Sensing System," *Proc. 2015 IEEE Int. Conf. Pervasive Comput. Commun. Work.*, p. (in-press), 2015.

[9] S. R. Safavian and D. Landgrebe, "A survey of decision tree classifier methodology," *IEEE Trans. Syst. Man. Cybern.*, vol. 21, no. 3, pp. 660–674, 1991.

[10] B. Scholkopft and K.-R. Mullert, "Fisher discriminant analysis with kernels," *Neural networks signal Process. IX*, vol. 1, p. 1, 1999.

[11] J. W. Riley, C. Alfons, E. Fredäng, and P. Lind, "Nearest Neighbor Classifiers," 2009.

[12] C. J. C. Burges, "A Tutorial on Support Vector Machines for Pattern Recognition," *Data Min. Knowl. Discov.*, vol. 2, no. 2, pp. 121–167, Jun. 1998.